\documentclass[runningheads]{llncs}
\usepackage[T1]{fontenc}
\usepackage{graphicx}
\usepackage{subcaption}
\usepackage[hidelinks]{hyperref} 
\usepackage{color}

\usepackage{float}
\usepackage{multicol}
\usepackage{array}
\usepackage{multirow}
\usepackage{amsmath}
\usepackage{lipsum}
\usepackage{dblfloatfix}
\usepackage{amsfonts}
\usepackage{amssymb}
\usepackage{tabularx}
\usepackage{orcidlink}
\usepackage{enumitem}
\setlist{nosep}

\newcommand{\repthanks}[1]{\textsuperscript{\ref{#1}}}
\makeatletter
\patchcmd{\maketitle}
{\def\thanks}
{\let\repthanks\repthanksunskip\def\thanks}
{}{}
\patchcmd{\@maketitle}
{\def\thanks}
{\let\repthanks\@gobble\def\thanks}
{}{}
\newcommand\repthanksunskip[1]{\unskip{}}
\makeatother

\begin{document}
	\title{MythraGen: Two-Stage Retrieval Augmented Art Generation Framework}
	\titlerunning{Retrieval Augmented Art Generation}

		\author{Quang-Khai Le \thanks{These authors contributed equally to this research. \protect\label{X}}\inst{1,2}\orcidlink{0009-0004-3733-6496}
		\and
		Cong-Long Nguyen \repthanks{X}\inst{1,2}\orcidlink{0009-0000-7286-8691}
		\and
		Minh-Triet Tran \inst{1,2}\orcidlink{0000-0003-3046-3041}
		\and
		Trung-Nghia Le \thanks{Corresponding author.} \inst{1,2}\orcidlink{0000-0002-7363-2610}}
	\authorrunning{Q.K. Le et al.}

	\institute{University of Science, Ho Chi Minh city, Vietnam
 \and Vietnam National University, Ho Chi Minh city, Vietnam\\
		\email{\{22120148,22120191\}@student.hcmus.edu.vn}, 
		\email{\{tmtriet,ltnghia\}@fit.hcmus.edu.vn}
		}
	\maketitle
 
 \begin{figure}[H]
  \includegraphics[width=\textwidth, trim=0pt 130pt 0pt 150pt, clip]{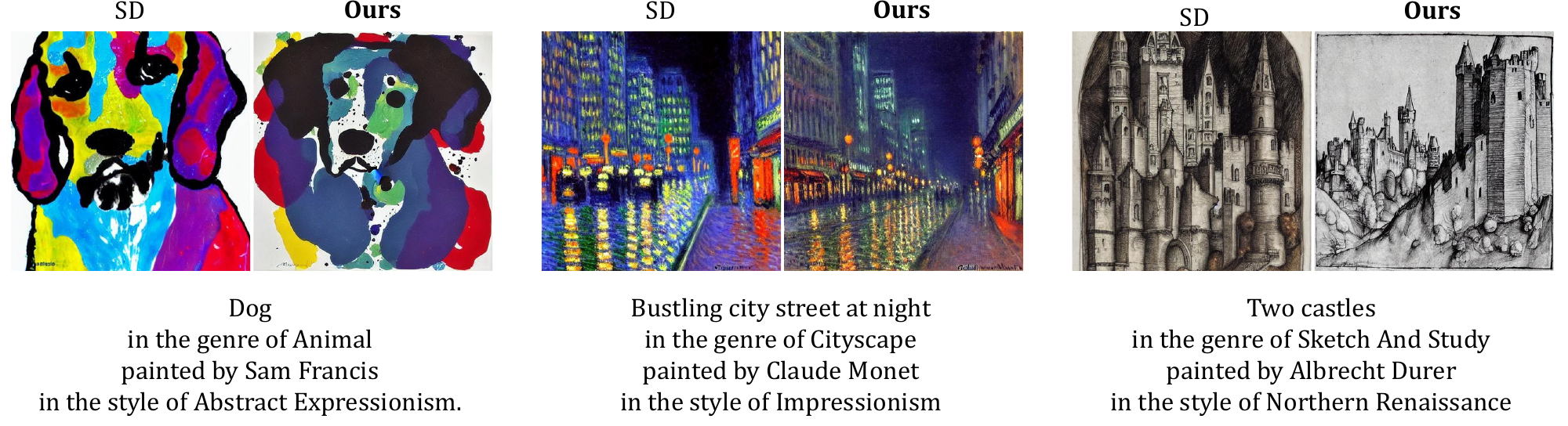}
  \vspace{-5mm}
  \caption{Examples of images generated by our method against Stable Diffusion.}
  \label{wrong_generation}
\end{figure}

 \begin{abstract}
 Text-to-image generation has seen rapid advancements, especially with the development of generative models. However, challenges remain in achieving high-quality, contextually accurate image outputs that faithfully match the provided textual descriptions, especially in artistic generation. In this paper, we present a simple yet efficient retrieval augmented generation framework, namely MythraGen, for text-to-artistic image generation by integrating an art retrieval mechanism with LoRA-based model fine-tuning. Our method extracts features from a large-scale art dataset, optimizing the generation process by combining artist-specific styles and content. Particularly, retrieved images from an external art database that have the highest similarity to the query prompt are used to finetune Stable Diffusion using LoRA for desired art generation. Experimental results and user studies on the WikiArt dataset show that our proposed method can generate artworks that closely match the user’s input, significantly outperforming existing solutions.
\end{abstract}

\section{Introduction}
    In art analysis, content and style are two fundamental elements. Content describes the concepts depicted in the image, such as objects, people, or locations. Style, on the other hand, describes the visual appearance of the artwork, including its color, composition, and shape. Furthermore, each artist expresses his/her own style, creating unique features in his/her works. Through the unique combination of content and the artist's individual style that makes each piece of art special.

Recent advances in deep learning have facilitated powerful breakthroughs in text-to-image generation (T2I)\cite{NEURIPS2021_a4d92e2c,10.1145/3652583.3658071,10.1145/3618322,Rombach_2022_CVPR,10.5555/3600270.3602913,9879397}. T2I methods can incorporate a specific style into generated images by taking a textual description of the style as a prompt. However, conveying artistic style through text descriptions has limitations. These descriptions are often less expressive and informative than visual representations of style, so the style features of T2I outputs are often rough and lack details. 

Recent fine-tuning techniques such as Dream-booth \cite{ruiz2022dreambooth}, Textual Inversion \cite{gal2022image}, and Low-Rank Adaptation (LoRA) \cite{hu2022lora} can enhance adaptability to specific T2I generation tasks, and show convincing capability in creating images with unique content and style. Among these methods, LoRA stands out and gains extensive adoption among art designers and T2I enthusiasts, due to its advantages of low-cost an computational efficiency, making it user-friendly and suitable for consumer devices. However, an artist can paint in many different styles. When extending this to hundreds of artists, the need to fine-tune or retrain the model for each artist's style becomes impractical. This process requires a vast amount of computational resources and time, making these methods unrealistic for large-scale application. 

To address these issues, we propose MythraGen, a simple yet efficient retrieval augmented art generation framework. First, we employ a retrieval technique to search for paintings from an external database that have the highest similarity in content, genre, and style to the artist described in the prompt. Our art retrieval technique leverages BLIP-2 to encode both the visual features of each image and its associated metadata, including captions, genre, style, and artist information. These encoded features are combined into a comprehensive feature vector, which is indexed using FAISS~\cite{douze2024faiss} to facilitate the retrieval of relevant images. Then, we utilize LoRA algorithm \cite{hu2022lora} for finetuning Stable Diffusion on the identified paintings. This two-stage framework allows for the flexible combination of different styles from various artists and content, optimizing image generation while ensuring the quality of the created images.

In this paper, the WikiArt dataset \cite{artgan2018}, consisting of 80,000 unique images from more than 1,100 artists across 27 styles, was used as the external art database. We also leveraged a zero-shot classifier based on Visual Question Answering (VQA) to annotate genres for round 16,000 images missing labels. Extensive experiments and user study showcase an impressive performance our MythraGen, outperforming existing existing open sources and commercial image generation methods in all evaluation metrics.

Our contributions can be summarized as follows: 
\begin{itemize}
    \item We introduce a simple yet retrieval augmented art generation framework called MythraGen to create artistic images with desired content, genre, and style from the query prompt.
    \item We train LoRA models on various artistic styles and then mix them to create superior composite results.
    \item We employ Visual Question Answering (VQA) based zero-shot classification to automatically label the genre for 16,452 unclassified images in the WikiArt dataset.
    \item We conduct a comprehensive evaluation of our proposed framework in reproducing different artistic styles. Experimental results and user study demonstrate the superiority of our proposed method against existing open sources and commercial image generation methods.
\end{itemize}

\section{Related Work}

\subsection{Art Retrieval}

Similarity search algorithms have played a crucial role in many artificial intelligence applications, with K-Nearest Neighbors (KNN) and Approximate Nearest Neighbors (ANN) being commonly used methods. KNN works by finding the closest points in the dataset to a query point based on a specified distance metric, making it useful for classification and regression tasks. ANN, on the other hand, provides faster search results by approximating the nearest neighbors, making it suitable for high-dimensional data where exact searches are computationally expensive. 

Recently, the Facebook AI Similarity Search (FAISS) library \cite{douze2024faiss} is a powerful tool for efficient similarity search and clustering of high-dimensional data, enabling developers to quickly find similar items in large datasets. It is particularly useful for tasks such as image retrieval, recommendation systems, and natural language processing, where finding similar items in large datasets is crucial. Therefore, this paper utilized FAISS due to its ability to quickly and accurately search for similar embedding vectors in latent space. This is especially useful when working with large datasets like WikiArt (around 80k images), allowing us to rapidly gather relevant images to support image generation processes. 


To ensure accurate retrieval, the alignment of image and text embeddings is crucial for effective cooperation. Inspired by the vision-language pre-training model BLIP-2 \cite{Li2023BLIP2BL}, which produces high-quality text-aligned visual representations, we adapted it to extract text-aligned subject representations, improving the model's ability to understand and generate content across modalities.

By integrating BLIP-2 \cite{Li2023BLIP2BL} with FAISS, we harness the strengths of advanced vision-language pre-training models and efficient similarity search algorithms. This combination allows us to improve the accuracy of image-text retrieval, providing a more precise and comprehensive dataset for further applications. 

\begin{figure}[t!]
    \centering
    \includegraphics[width=\linewidth, trim=0pt 110pt 0pt 0pt, clip]{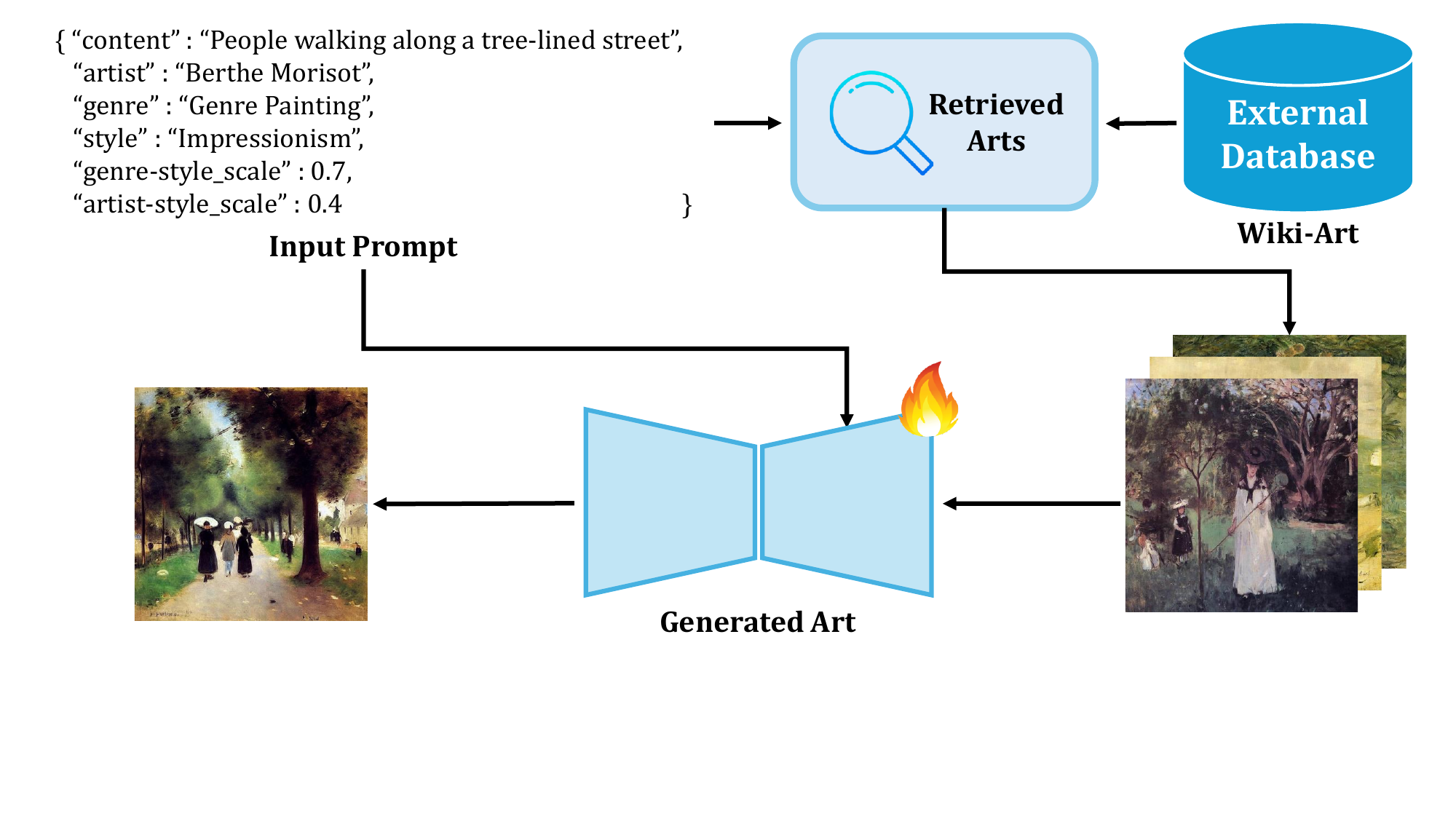}
    \caption{Overview of the proposed MythraGen framework, with two main stages: (a) the Art Retrieval retrieves images to enhance the image generation process and (b) the Art Generation generates images based on the user's input text combined with the images provided by the Art Retrieval module.}
    \label{Overview_method}
    \vspace{-5mm}
\end{figure}

\subsection{Art Generation}

Early text-to-image models \cite{pmlr-v48-reed16,Xu_2018_CVPR,8411144,Zhu2019DMGANDM} made significant progress by utilizing Generative Adversarial Networks (GANs) \cite{gal2022image} trained on large paired image-caption datasets, which can lead to model collapse issues \cite{Brock2018LargeSG,NEURIPS2021_49ad23d1,JMLR:v23:21-0635}. Recently, diffusion models \cite{Ho2020DenoisingDP,Rombach2021HighResolutionIS,nichol2022glide,10.5555/3600270.3602913} have become powerful in text-to-image (T2I) tasks due to their ability to generate high-quality images and their flexibility in adapting images to the context of the text. GLIDE \cite{nichol2022glide} and Imagen \cite{saharia2022photorealistic} employ classifier-free guidance by replacing the label in the class-conditioned diffusion model with text descriptions of the images. Stable Diffusion \cite{rombach2022high} utilizes VQ-GAN for the latent representation of images, enhancing photorealism through an adversarial objective. DALL-E2 \cite{ramesh2022hierarchical} employs a multimodal latent space where image and text embeddings are aligned to generate images reflecting a deeper level of language understanding. However, these models often struggle when handling prompts containing less common genres or styles related to artists. Instead of generating a suitable image, they tend to either create nonexistent genres or styles or use similar but more popular ones, which does not align with the user's intent, leading to a mismatch between the original prompt and the final image.

Unifying both image retrieval and generation processes, Re-Imagen \cite{Chen2022ReImagenRT} addressed augmenting rare entities to improve image quality. However, our goal is to use retrieval to enhance the less common drawing styles of various artists, while ensuring that when these styles are applied, the main content of the prompt remains preserved. Additionally, Re-Imagen trains its image generation process on the cascaded diffusion model \cite{JMLR:v23:21-0635}, while our method uses LoRA \cite{hu2022lora} to fine-tune the Stable Diffusion model \cite{Li2023BLIP2BL} for reducing required resources and speed up the training process.



\begin{figure}[t!]
    \centering
  \includegraphics[width=\linewidth, trim=0pt 230pt 0pt 0pt, clip]{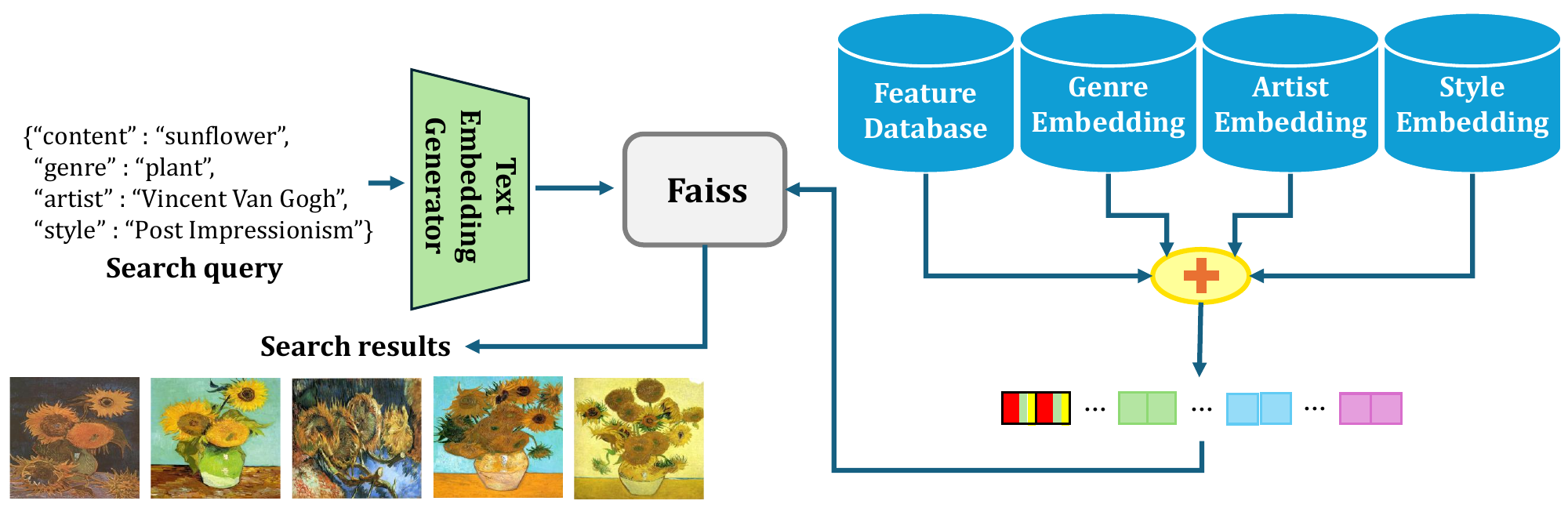}
  \caption{Pipeline of the Art Retrieval module.}
  \label{Art_retrieval}
  \vspace{-5mm}
\end{figure}

\section{Proposed Method}
\subsection{Overview}


Figure~\ref{Overview_method} illustrates an overview of our proposed two-stage retrieval augmented art generation framework. The input prompt is fed into the Art Retrieval module to search for images that are similar to the input text content. In the Art Generation module, we use the images obtained from the Art Retrieval module to enrich the data during the image generation process, which is also controlled by the text embedding vector. Finally, the Art Generation module produces a target image that has the highest similarity to the user's input text.

\subsection{Art Retrieval}
In this section, we present our approach for retrieval from an external database to find images with the highest similarity in content, genre, and style to the artist described in the input query. Figure~\ref{Art_retrieval} provides a visual representation of the Art Retrieval module, built around the Bootstrapping Language Image Pre-training (BLIP-2 \cite{Li2023BLIP2BL}) architecture. First, each 256-dimensional vector from the feature, genre, artist, and style databases, which are related to each other, are concatenated into a single 1024-dimensional vector. This vector is then fed to the FAISS system for indexing to support the retrieval process. When the input query is processed by the Text Embedding Generator, generated by the Q-Former component of BLIP-2 \cite{Li2023BLIP2BL}, the resulting vector is sent to the FAISS system to return a set of images with the highest similarity in content, genre, and style to the artist described in the input query.

\begin{figure}[t!]
\centering \includegraphics[width=0.8\linewidth]{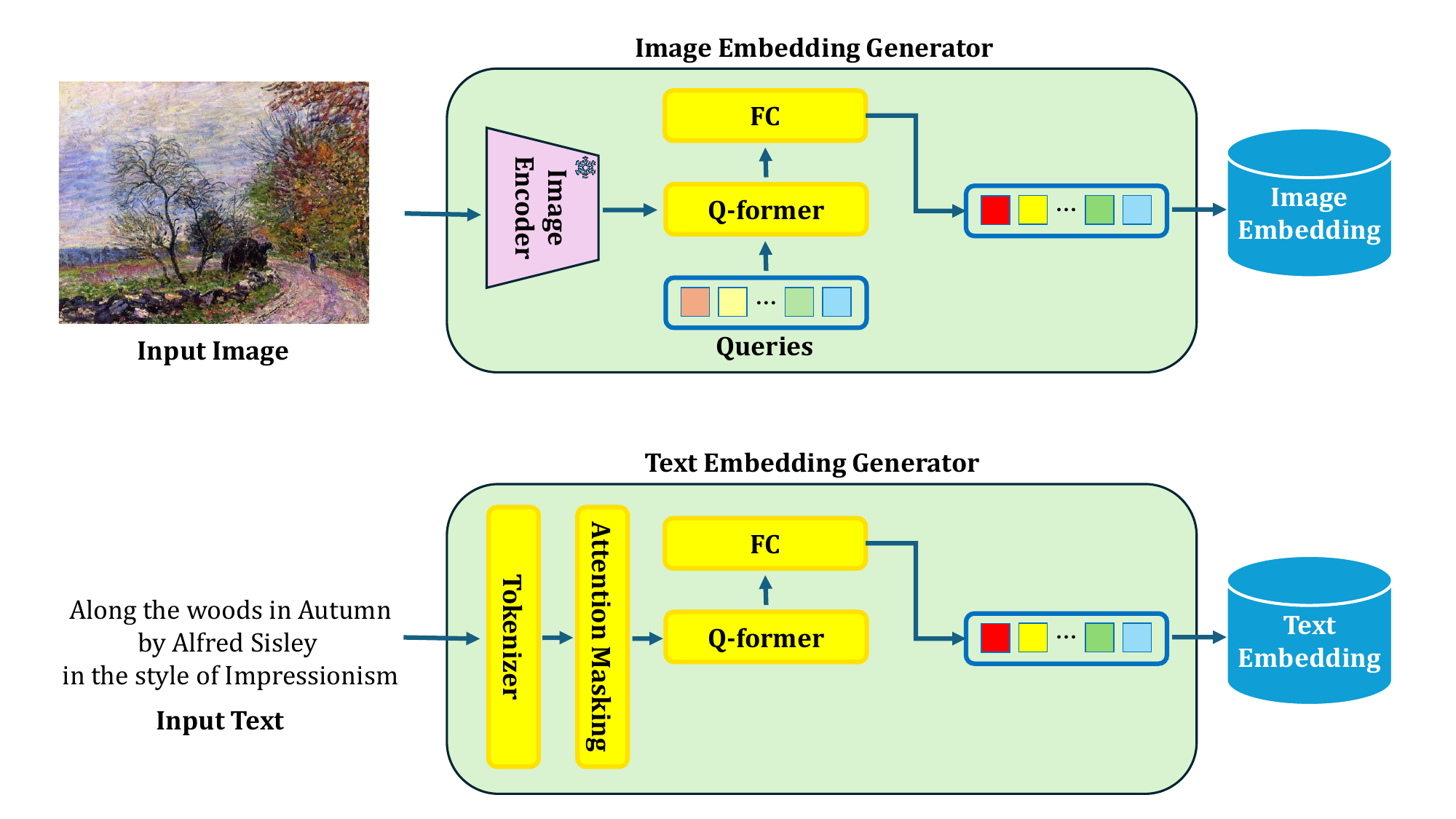}
  \caption{Multimodal representation, where image and text embeddings processes rely primarily on the Q-former.}
  \label{Art_encoder}
  \vspace{-5mm}
\end{figure}


The Image Embedding Generator consists of two main components: a frozen pre-trained image encoder and a multimodal encoder called Q-Former (see Fig.~\ref{Art_encoder}). The process begins with the input image, which is passed through the image encoder to extract visual features. These features are then combined with learnable query tokens and fed into the Q-Former. The output of the Q-Former is then passed through a fully connected layer and normalized to produce the final image feature vector. Similarly, the Text Embedding Generator tokenizes the input text and processes it through the Q-Former. The output from the Q-Former is then passed through a fully connected layer and normalized to produce the final text feature vector.




To improve the retrieval performance, we combine image embedding with caption and genre embeddings. Let $E_i$ with $i \in \{image, caption, genre\}$ represents the image, caption, and genre embeddings. We compute the weighted sum of the embeddings as follows:
\begin{equation}
\label{eq:sum}
    V =  \sum_{i} W_i * E_i,
\end{equation}
where $W_i$ denotes the corresponding weight of embedding $E_i$. The resulting vector $V$ becomes the final feature vector used for retrieval. 
By combining information from images, captions, and genres, a stronger feature vector is created for better retrieval performance for genre.

The feature vector may not perform well in retrieving style and artist information because it mainly captures the content and genre of the artwork. Moreover, many art styles share significant similarities in technique, such as \textit{Action Painting} and \textit{Abstract Expressionism}, or between movements like \textit{Cubism}, \textit{Analytical Cubism} and \textit{Synthetic Cubism} or \textit{Impressionism} and \textit{Post-Impressionism}, etc. This makes it difficult to distinguish between closely related styles or specific artists. To address this, we concatenate additional style embedding and artist embedding to enrich the data for the vector before the indexing phase (See Fig.~\ref{Art_retrieval}). By incorporating these additional embeddings, the model can better capture the nuances of style and artist characteristics, thereby improving retrieval performance in these specific areas. Moreover, despite the increased complexity from the added embeddings, our optimized retrieval system maintain an impressive speed, achieving a retrieval time of just $0.1$ second per query. This efficiency demonstrates the robustness of our approach, ensuring that even with enriched data, the system remains highly responsive.

\subsection{Art Generation}

\begin{figure}[t!]
    \centering
  \includegraphics[width=\linewidth]{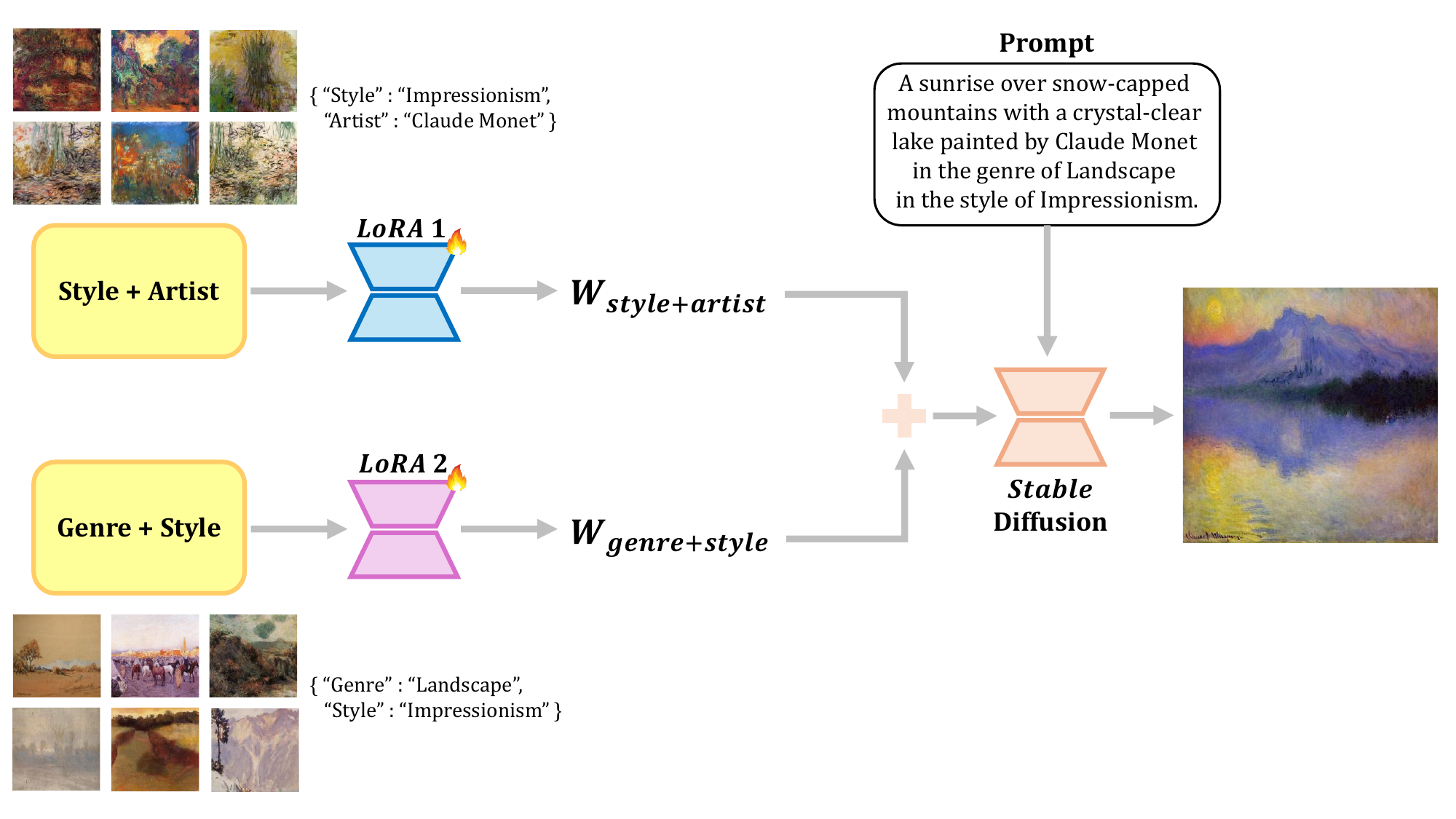}
  \caption{LoRA combination in Art Generation module. The first LoRA model is fine-tuned for the artist and style ($W_{style+artist}$), while the second LoRA model is fine-tuned for the genre and style ($W_{genre+style}$). Both are combined and applied to the Stable Diffusion model to generate an image that faithfully reflects the input prompt, incorporating the specific style, artist, and genre.}
  \label{Art_generation}
  \vspace{-5mm}
\end{figure}

We utilize the LoRA technique \cite{hu2022lora} to fine-tune two distinct LoRA models, each is specifically trained on different types of input data sourced from the Art Retrieval module (see Fig.~\ref{Art_generation}). The fine-tuning process allows us to tailor each LoRA model to accurately reflect particular artistic attributes, such as genre, style, and artist characteristics. After fine-tuning, each LoRA model is assigned its own unique weight, carefully calibrated to balance the influence of each model on the final output. These weighted models are then seamlessly mixed together, in conjunction with the given prompt, to generate the final target image. This process allows to create highly nuanced and accurate representations of artistic styles. For example, as shown in Fig.~\ref{Art_generation}, by combining the \textit{Landscape genre} with the \textit{Impressionism style} and integrating the influence of \textit{the artist Claude Monet}, the technique produces an image of a sunrise over snow-capped mountains with a crystal-clear lake. The resulting image not only depicts the specified scene but also strongly embodies Monet's distinctive Impressionist style, showcasing the effectiveness of our approach in capturing complex artistic nuances.

\section{Experiments}
\subsection{Implementations}

In our experiments, we leveraged the PyTorch deep learning framework on a computer equipped with an NVIDIA RTX A4500 GPU with 24 GB. For the Art Retrieval module, we used BLIP-2 \cite{Li2023BLIP2BL} pre-trained with ViT-L/14 to embed the image and its related metadata. Weights in Eq. \ref{eq:sum} is set $W_{\text{image}} = 1.0$, $W_{\text{caption}} = 0.9$, and $W_{\text{genre}} = 0.75$. 

Regarding the Art Generation module, we employed Stable Diffusion V1.5 \cite{Rombach_2022_CVPR} as the backbone. We utilized two LoRA models for our experiments: one fine-tuned for genre + style and another fine-tuned for artist + style. For both LoRA, we set $ \text{max\_train\_steps} $ to 4095, using AdamW8bit as the optimizer to save memory, with xformers enabled for memory optimization and $ \text{mixed\_precision} = \text{"bf16"} $ for faster computation without compromising quality. The learning rates were set to $1 \times 10^{-4}$ for the U-Net and $5 \times 10^{-5}$ for the text encoder. The LoRA network was configured with $ \text{network\_dim} = 32 $ and $ \text{network\_alpha} = 1 $, adjusting the rank and scaling of the LoRA layers for efficient fine-tuning.

\subsection{Dataset}

\begin{figure}[t!]
    \centering
    \includegraphics[width=\linewidth, trim=0pt 180pt 0pt 0pt, clip]{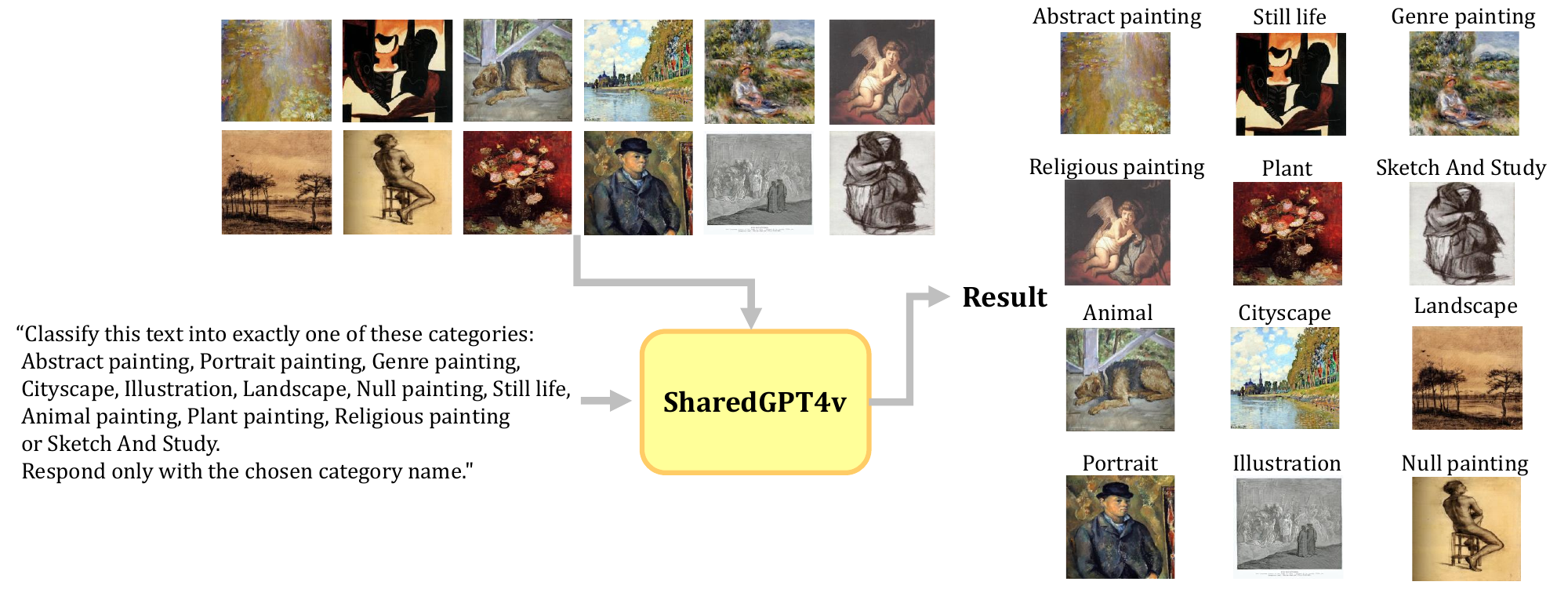}
    \caption{Visual Question Answering (VQA) model used to classify the genre of an image when the genre is unknown.}
    \label{genre_label}
\end{figure}

We used WikiArt as the external database for all experiments in this paper. WikiArt has complete labels for artists and styles, but over 16,452 images are missing genre labels. Therefore, we employed a Visual Question Answering (VQA) model (i.e., ShareGPT4V \cite{chen2023sharegpt4v}) to label these 16,452 images according to 12 corresponding genres (genre painting, illustration, landscape, etc.) as described in Fig.~\ref{genre_label}, where the model analyzes each image and assigns the appropriate genre label based on visual content.


We performed retrieval on the WikiArt dataset using three separate categories: artist, styles, and genres, as well as a combined category that includes all three.

\subsection{Art Retrieval}



\begin{table}[t!]
    \centering
    \caption{Results of different BLIP-2 versions on retrieval across characteristics.}
    \label{genre}
    \resizebox{\textwidth}{!}{%
    \begin{tabular}{|l|c|c|c|c|c|}
        \hline
        \multirow{2}{*}{\textbf{Model}} & \multicolumn{5}{c|}{\textbf{Genres}} \\ \cline{2-6}
        & \textbf{mAP@5} & \textbf{mAP@15} & \textbf{mAP@25} & \textbf{mAP@40} & \textbf{mAP@50} \\ \hline
        \raggedright BLIP2 using ViT-g/14 (EVA-CLIP) & 100\% & 98.4\% & 98.4\% & 98.6\% & 98.6\% \\ \hline
        \raggedright BLIP2 using ViT-L/14 (CLIP) & 100\% & \textbf{99.8\%} & \textbf{99.7\%} & \textbf{99.6\%} & \textbf{99.5\%} \\ \hline
        \raggedright BLIP2 finetuned on COCO \cite{cocodataset} & 100\% & 98.8\% & 98.3\% & 97.8\% & 97.7\% \\ \hline \hline
        
        \multirow{2}{*}{\textbf{Model}} & \multicolumn{5}{c|}{\textbf{Artists, Styles}} \\ \cline{2-6}
        & \textbf{mAP@5} & \textbf{mAP@15} & \textbf{mAP@25} & \textbf{mAP@40} & \textbf{mAP@50} \\ \hline
        \raggedright BLIP2 using ViT-g/14 (EVA-CLIP) & 100\% & 100\% & 100\% & 100\% & 100\% \\ \hline
        \raggedright BLIP2 using ViT-L/14 (CLIP) & 100\% & 100\% & 100\% & 100\% & 100\% \\ \hline
        \raggedright BLIP2 finetuned on COCO \cite{cocodataset} & 100\% & 100\% & 100\% & 100\% & 100\% \\ \hline \hline
        
        \multirow{2}{*}{\textbf{Model}} & \multicolumn{5}{c|}{\textbf{Genres $+$ Artists $+$ Styles}} \\ \cline{2-6}
        & \textbf{mAP@5} & \textbf{mAP@15} & \textbf{mAP@25} & \textbf{mAP@40} & \textbf{mAP@50} \\ \hline
        \raggedright BLIP2 using ViT-g/14 (EVA-CLIP) & 90.5\% & 92.1\% & 92.4\% & 92.4\% & 92.6\% \\ \hline
        \raggedright BLIP2 using ViT-L/14 (CLIP) & 92.8\% & \textbf{94.8\%} & \textbf{94.4\%} & \textbf{93.8\%} & \textbf{94.2\%} \\ \hline
        \raggedright BLIP2 finetuned on COCO \cite{cocodataset} & \textbf{93.4\%} & 94\% & 93.7\% & 93.7\% & 93.7\% \\ \hline
    \end{tabular}
    }
    \vspace{-5mm}
\end{table}

\begin{table}[t!]
    \centering
    \caption{Results of comparing the performance between using a standalone image embedding and combining it with caption and genre information across different versions of BLIP-2.}
    \label{featureVector}
    \resizebox{\textwidth}{!}{%
    \begin{tabular}{|l|c|c|c|c|c|}
        \hline
        \multirow{2}{*}{\textbf{Embedding vector}} & \multicolumn{5}{c|}{\textbf{BLIP2 using ViT-g/14 (EVA-CLIP)}} \\ \cline{2-6}
        & \textbf{mAP@5} & \textbf{mAP@15} & \textbf{mAP@25} & \textbf{mAP@40} & \textbf{mAP@50} \\ \hline
        Image embedding & 82.7\% & 78.5\% & 77.0\% & 75.7\% & 75.2\% \\ \hline
        Image $+$ Caption $+$ Genre embedding & \textbf{100\%} & \textbf{98.4\%} & \textbf{98.4\%} & \textbf{98.6\%} & \textbf{98.6\%} \\ \hline

        \multirow{2}{*}{\textbf{Embedding vector}} & \multicolumn{5}{c|}{\textbf{BLIP2 using ViT-L/14 (CLIP)}} \\ \cline{2-6}
        & \textbf{mAP@5} & \textbf{mAP@15} & \textbf{mAP@25} & \textbf{mAP@40} & \textbf{mAP@50} \\ \hline
        Image embedding & 89.0\% & 81.3\% & 78.6\% & 75.3\% & 74.1\% \\ \hline
        Image $+$ Caption $+$ Genre embedding & \textbf{100\%} & \textbf{99.8\%} & \textbf{99.7\%} & \textbf{99.6\%} & \textbf{99.5\%} \\ \hline

        \multirow{2}{*}{\textbf{Embedding vector}} & \multicolumn{5}{c|}{\textbf{BLIP2 finetuned on COCO \cite{cocodataset}}} \\ \cline{2-6}
        & \textbf{mAP@5} & \textbf{mAP@15} & \textbf{mAP@25} & \textbf{mAP@40} & \textbf{mAP@50} \\ \hline
        Image embedding & 84.6\% & 79.8\% & 77.0\% & 75.5\% & 74.9\% \\ \hline
        Image $+$ Caption $+$ Genre embedding & \textbf{100\%} & \textbf{98.8\%} & \textbf{98.3\%} & \textbf{97.8\%} & \textbf{97.7\%}\\ \hline   
    \end{tabular}
    }
    \vspace{-5mm}
\end{table}

Table \ref{genre} compares the performance of three BLIP-2 versions in retrieval tasks across genres, artists, and styles. All models get a perfect mAP@5 of 1.0, showing high accuracy for the top 5 results. For genres, BLIP2 with ViT-L/14 (CLIP) scores the highest mAP@50 at 99.5\%, followed by ViT-g/14 (EVA-CLIP) at 98.6\%, and the COCO-finetuned model at 97.7\%. In artist-based and style-based retrieval, all models reach a perfect mAP of 1.0 at all levels, showing equal skill in retrieving artist and style information. However, when combining genre, artist, and style , the COCO-finetuned model performs best at mAP@5 with 93.4\%, but ViT-L/14 (CLIP) performs better as the number of queries increases, reaching the highest mAP@50 at 94.2\%. Overall, ViT-L/14 (CLIP) performs best in more complex tasks, while ViT-g/14 (EVA-CLIP) has slightly lower results.

Table \ref{featureVector} shows the comparison between using only image embedding and combining it with caption and genre information across different versions of BLIP-2. The results use metrics like mAP@5, mAP@15, mAP@25, mAP@40, and mAP@50 to measure accuracy. Using only image embedding gives good results, such as 82.7\% with BLIP-2 using ViT-g/14 (EVA-CLIP) and 89.0\% with BLIP-2 using ViT-L/14 (CLIP) at mAP@5. However, when adding caption and genre embeddings, the performance improves significantly, reaching over 98\% for most cases.

\subsection{Art Generation}

\begin{table}[t!]
    \centering
    \caption{Comparison of methods in terms of CLIP-T, CLIP-I, and FID metrics.}
    \label{compareMethods}
    \begin{tabularx}{\textwidth}{|l|>{\centering\arraybackslash}X|>{\centering\arraybackslash}X|>{\centering\arraybackslash}X|>{\centering\arraybackslash}X|>{\centering\arraybackslash}X|}
        \hline
        \multirow{2}{*}{\textbf{Metrics}} & \multicolumn{4}{c|}{\textbf{Methods}} \\ \cline{2-5}
         & \textbf{MythraGen} & \textbf{BingAI} & \textbf{Midjourney} & \textbf{SD} \\ \hline
         CLIP-T $\uparrow$ & \textbf{30.68} & 27.77  & 30.13 & 29.61  \\ \hline
         CLIP-I $\uparrow$ & \textbf{79.84} & 66.82 & 65.19 & 75.29  \\ \hline
         FID $\downarrow$ & \textbf{322.9} & 373.85 & 329.22 & 325.79  \\ \hline
    \end{tabularx}
\end{table}



\subsubsection{Compared Methods. }
\label{sec:exp_methods}
We compared our method with state-of-the-art (SOTA) methods, including Stable Diffusion (SD) version 2.0 \cite{Rombach_2022_CVPR} and two well-known commercial products: BingAI \footnote{\url{https://www.bing.com/images/create/}} and Midjourney \footnote{\url{https://www.midjourneyfree.ai/}}. 

\subsubsection{Dataset. }
\label{sec:exp_dataset}
The comparison was conducted on a customized dataset extracted from WikiArt. The dataset consists of 50 randomly selected (genre, artist, style) combinations from WikiArt, along with 50 prompts based on scenes that we generated. Each prompt was carefully crafted to evaluate the methods' ability to accurately generate images that adhere to both the style and content described. This customized dataset allowed for a comprehensive comparison of the models' performance in generating stylistically consistent and content-accurate images. 

\subsubsection{Experimental Results. }

Table \ref{compareMethods} shows that our MythraGen outperforms the other methods across all three metrics. In terms of CLIP-T, which measures the textual similarity between the prompt and the generated image, MythraGen achieves the highest score at 30.68, surpassing Midjourney (30.13), SD (29.61), and BingAI (27.77). For CLIP-I, which measures the style similarity, MythraGen also lead with a score of 79.84, followed by SD (75.29), BingAI (66.82), and Midjourney (65.19). Finally, for FID, which measures the quality of the images, MythraGen obtains the lowest score (indicating better performance) at 322.9, compared to SD (325.79), Midjourney (329.22), and BingAI (373.85).

Fig.~\ref{compareMethods} illustrates the visual results of these methods. While SD \cite{Rombach_2022_CVPR} struggles to balance style and content due to poor representation of the text extracted from the reference image, the content of the images generated by BingAI, and Midjourney is relatively closer to the prompt, but their style differs from the style reference. In contrast, our method produces images that are more faithful to the style of the reference image, especially regarding brushstrokes, lines, etc. This demonstrates that our MythraGen method achieves a better balance between content similarity, style similarity, and generated quality according to objective metrics.

 \begin{figure}[t!]
  \includegraphics[width=\textwidth, trim=0pt 150pt 0pt 0pt, clip]{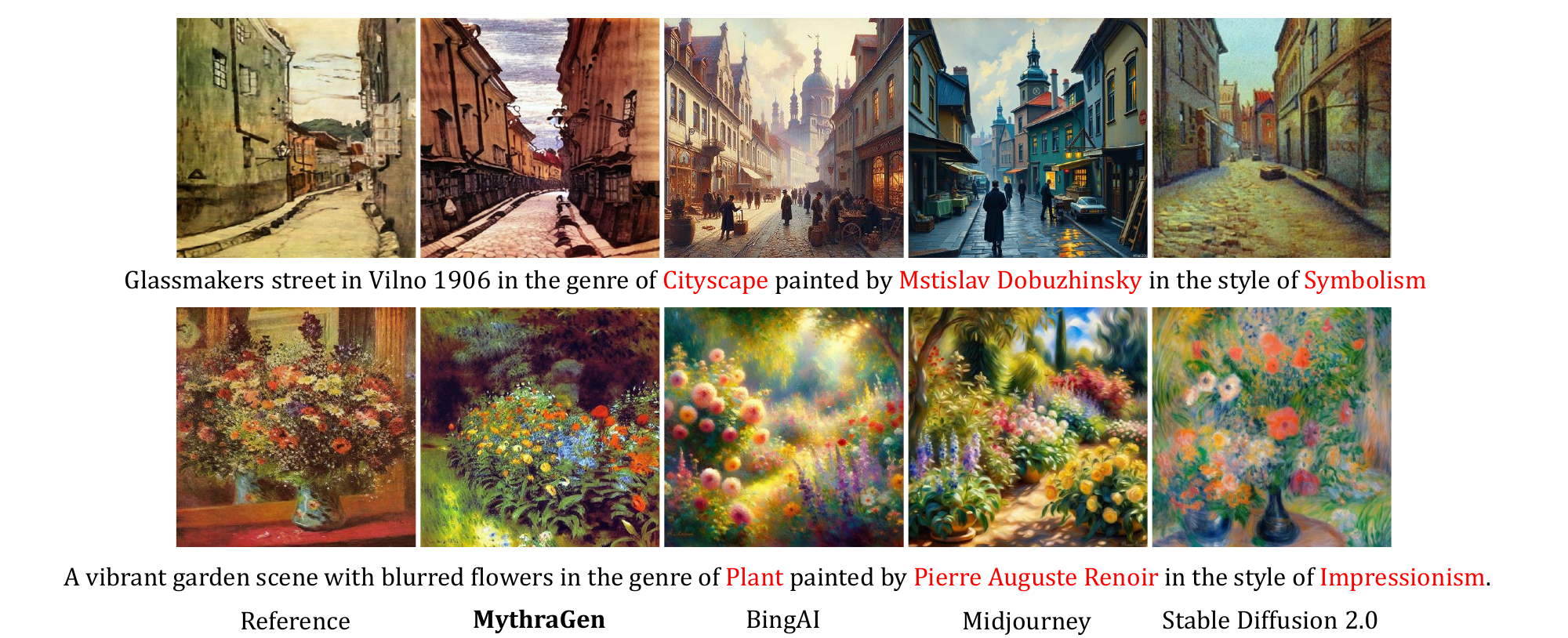}
  \caption{Qualitative comparison with SOTA methods based on style reference image. Methods like BingAI, Midjourney, and SD 2.0 lack specific stylistic information drawn from the original artist, leading to difficulties in balancing content and style. In contrast, MythraGen performs better in generating both style and content as intended.}
  \label{compareMethods}
  \vspace{-5mm}
\end{figure}


\subsection{Human Evaluation}





\subsubsection{Setup. }
We also conducted a human evaluation of MythraGen against three SOTA methods as in Section \ref{sec:exp_methods}, on the same dataset as in Section \ref{sec:exp_dataset}. The user study focused on two key criteria:
\begin{itemize}
    \item \textbf{Faithfulness} assesses the extent to which the generated image is true to both the entity's appearance and the text description. It considers factors such as object fidelity and text relevance, ensuring that the image accurately represents the caption provided 
    \item \textbf{Naturalness} evaluates the visual quality of the generated image. It encompasses aspects of style similarity to the original image, ensuring that the image adheres faithfully to the style described in the prompt without noticeable artifacts or discrepancies that detract from its overall aesthetic appeal.
\end{itemize}



The user study involves 32 participants (48.4\% of whom are male) with age between 11 and 60 (most of them are from 11 to 20). Each participant was asked to score the outputs from different methods on a scale of 1 (worst) to 5 (best) based on two primary criteria. Each participant evaluated a total of 30 images per method. 

\begin{figure}[t!]
    \centering
    \begin{subfigure}[t]{0.5\textwidth}
        \centering
        \includegraphics[trim=2pt 2pt 2pt 2pt, clip, height=1.2in]{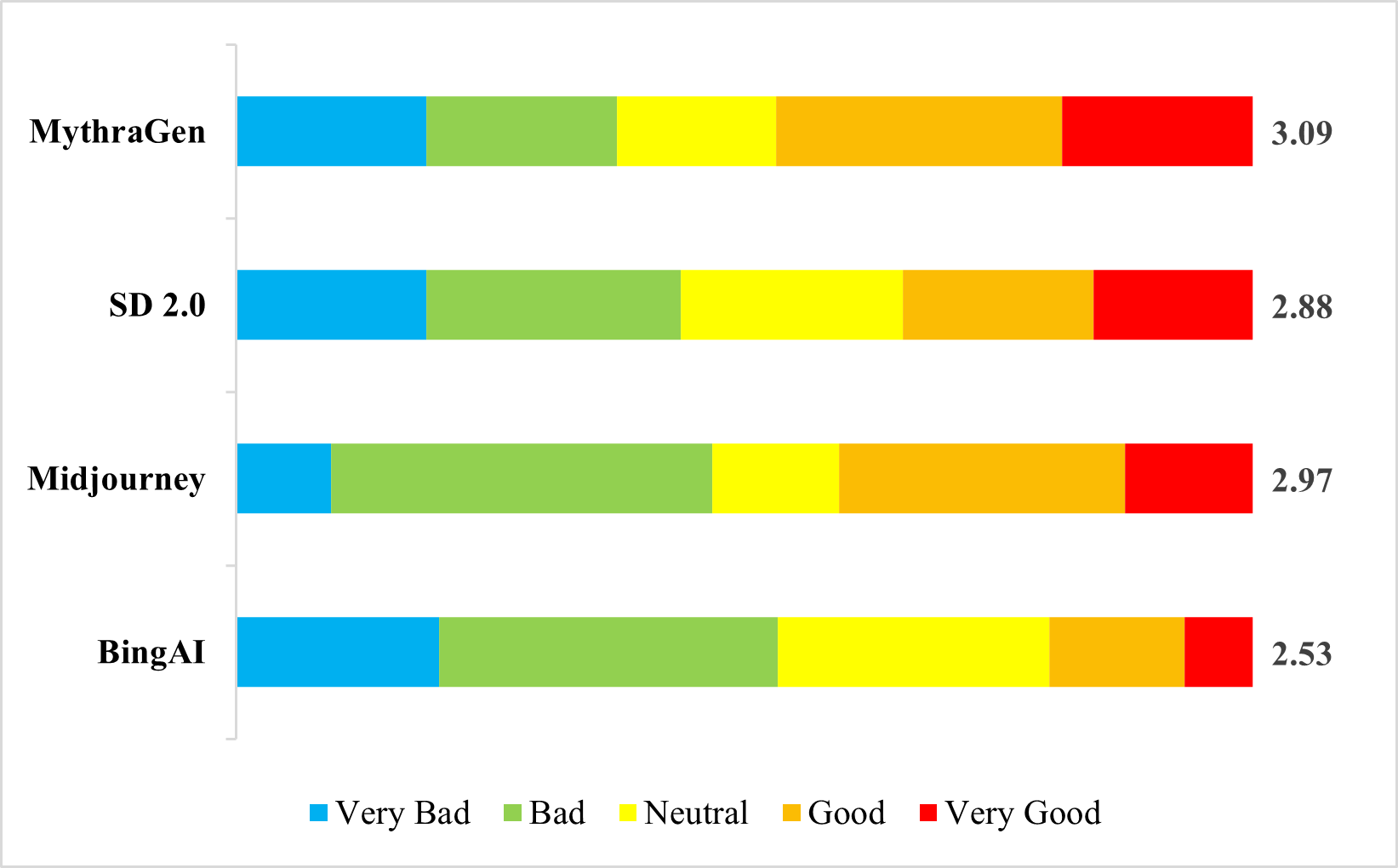}
        \caption{Faithfulness}
    \end{subfigure}%
    ~ 
    \begin{subfigure}[t]{0.5\textwidth}
        \centering
        \includegraphics[trim=2pt 2pt 2pt 2pt, clip, height=1.2in]{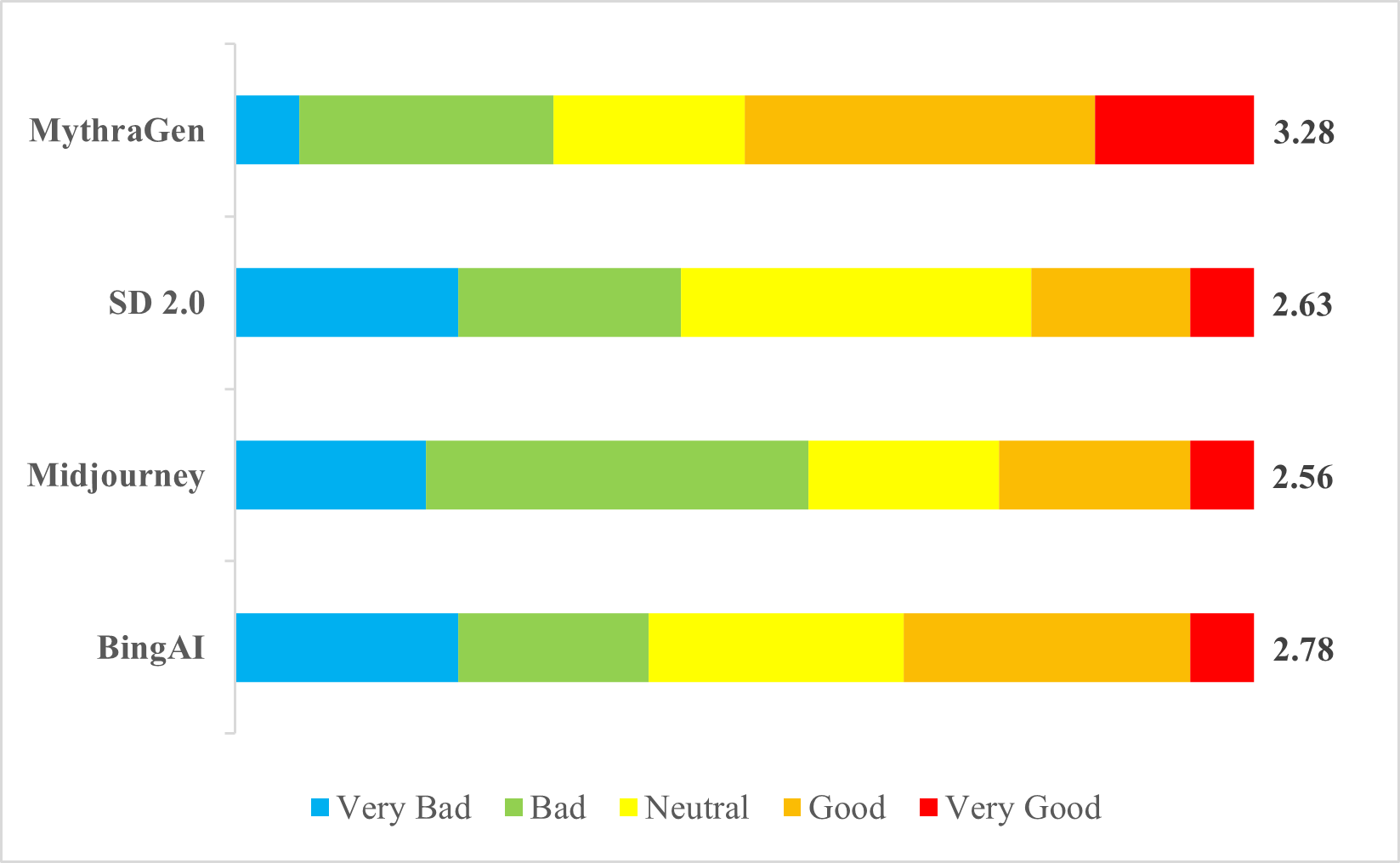}
        \caption{Naturalness}
    \end{subfigure}
    \caption{Humans evaluate the methods based on two criteria: Faithfulness and Naturalness.}
    \label{HumanEvaluation_pic}
\end{figure}

\subsubsection{Quantitative Results. }
Fig.~\ref{HumanEvaluation_pic} shows the quantitative results of the human evaluation, where MythraGen achieves the highest scores for both faithfulness and naturalness. For faithfulness, MythraGen gets an average score of 3.09, clearly improving over SD 2.0 (2.88), Midjourney (2.97), and BingAI (2.53). This result shows that MythraGen is better at generating images that accurately match the descriptions, both visually and textually. Similarly, for naturalness, MythraGen also outperforms other methods with an average score of 3.28. Participants find MythraGen's images more visually appealing and better aligned with the described styles in the input prompt. These results demonstrate that MythraGen not only excels in generating images that are faithful to their descriptions but also produces outputs that accurately reflect the requested artistic style, proving its effectiveness over SOTA methods in both accuracy and quality.





Our experimental results show that MythraGen is highly effective at generating images that faithfully represent both the text prompt and the desired artistic style. By using a retrieval-augmented approach, MythraGen leverages existing artwork to fine-tune the generation process, producing high-quality outputs that closely match user expectations. Additionally, the use of LoRA for efficient fine-tuning helps reduce computational costs while maintaining impressive performance, making the model accessible even on less powerful hardware.



\section{Conclusion}
In this paper, we present MythraGen, which represents the advancement in the field of text-to-artistic image generation by effectively combining retrieval-based techniques with LoRA fine-tuning. Our work not only enhances the quality and contextual accuracy of generated artworks but also enables the incorporation of diverse artistic styles that meet the user's expectations. Experimental results demonstrate that MythraGen outperforms existing methods in generating images that faithfully reflect text descriptions and are highly natural, as evidenced by user studies. We further demonstrate that our model is particularly effective at generating images from text that requires a greater diversity of artistic genres and periods. We believe that our work can inspire further innovations in the intersection of art and artificial intelligence, fostering deeper engagement with both creators and audiences. 

\textbf{Acknowledgement. }  This research is supported by research funding from Faculty of Information Technology, University of Science, Vietnam National University - Ho Chi Minh City.



\bibliographystyle{splncs04}
\bibliography{references}

\end{document}